\pdfoutput=1

\documentclass[11pt]{article}

\usepackage[final]{acl}

\usepackage{times}
\usepackage{latexsym}
\usepackage{amsmath}

\usepackage[T1]{fontenc}

\usepackage[utf8]{inputenc}

\usepackage{microtype}
\usepackage{newtxtext}

\usepackage{inconsolata}

\usepackage{graphicx}

\usepackage{tcolorbox}
\usepackage{hyperref}
\usepackage{url}
\usepackage{graphicx}
\usepackage{array}
\usepackage[bottom]{footmisc}
\usepackage{multicol}
\usepackage{pdfpages}

\title{MDBench: A Synthetic Multi-Document Reasoning Benchmark Generated with Knowledge Guidance}

\author{
Joseph J. Peper$^1$ \quad Wenzhao Qiu$^1$ \quad Ali Payani$^2$ \quad \textbf{Lu Wang}$^1$ \\ \\
$^1$Computer Science and Engineering, University of Michigan, Ann Arbor, MI \\
$^2$Cisco Research, San Jose, CA \\
\texttt{\{jpeper,qwzhao,wangluxy\}@umich.edu} \\
\texttt{apayani@cisco.com}
}

\newcommand\MDName{\textsc{MDBench}}

\usepackage{tcolorbox}
\usepackage{hyperref}
\usepackage{url}
\usepackage{graphicx}
\usepackage{array}
\usepackage[bottom]{footmisc}

\usepackage{amssymb}
\usepackage{pifont}

\begin{document}

\maketitle

\begin{abstract}
Natural language processing evaluation has made significant progress, largely driven by the proliferation of powerful large language models (LLMs). 
New evaluation benchmarks are of increasing priority as the reasoning capabilities of LLMs are expanding at a rapid pace. In particular, while \textit{multi-document} (MD) reasoning is an area of extreme relevance given LLM capabilities in handling longer-context inputs, few benchmarks exist to rigorously examine model behavior in this setting. Moreover, the multi-document setting is historically challenging for benchmark creation due to the expensive cost of annotating long inputs. \\
In this work, we introduce \textbf{\MDName{}}, a new dataset for evaluating LLMs on the task of multi-document reasoning. Notably, \MDName{} is created through a novel synthetic generation process, allowing us to \textit{controllably and efficiently generate challenging document sets} and the corresponding question-answer (QA) examples. Our novel technique operates on condensed structured seed knowledge, modifying it through LLM-assisted edits to induce MD-specific reasoning challenges. We then convert this structured knowledge into a natural text surface form, generating a document set and corresponding QA example. 
We analyze the behavior of frontier LLMs and prompting techniques, finding that \MDName{} poses significant challenges for all methods, even with relatively short document sets. We also see our knowledge-guided generation technique (1) allows us to readily perform targeted analysis of MD-specific reasoning capabilities and (2) can be adapted quickly to account for new challenges and future modeling improvements.
\end{abstract}

\begin{figure*}[t]
\centering
     \includegraphics[width=\textwidth]{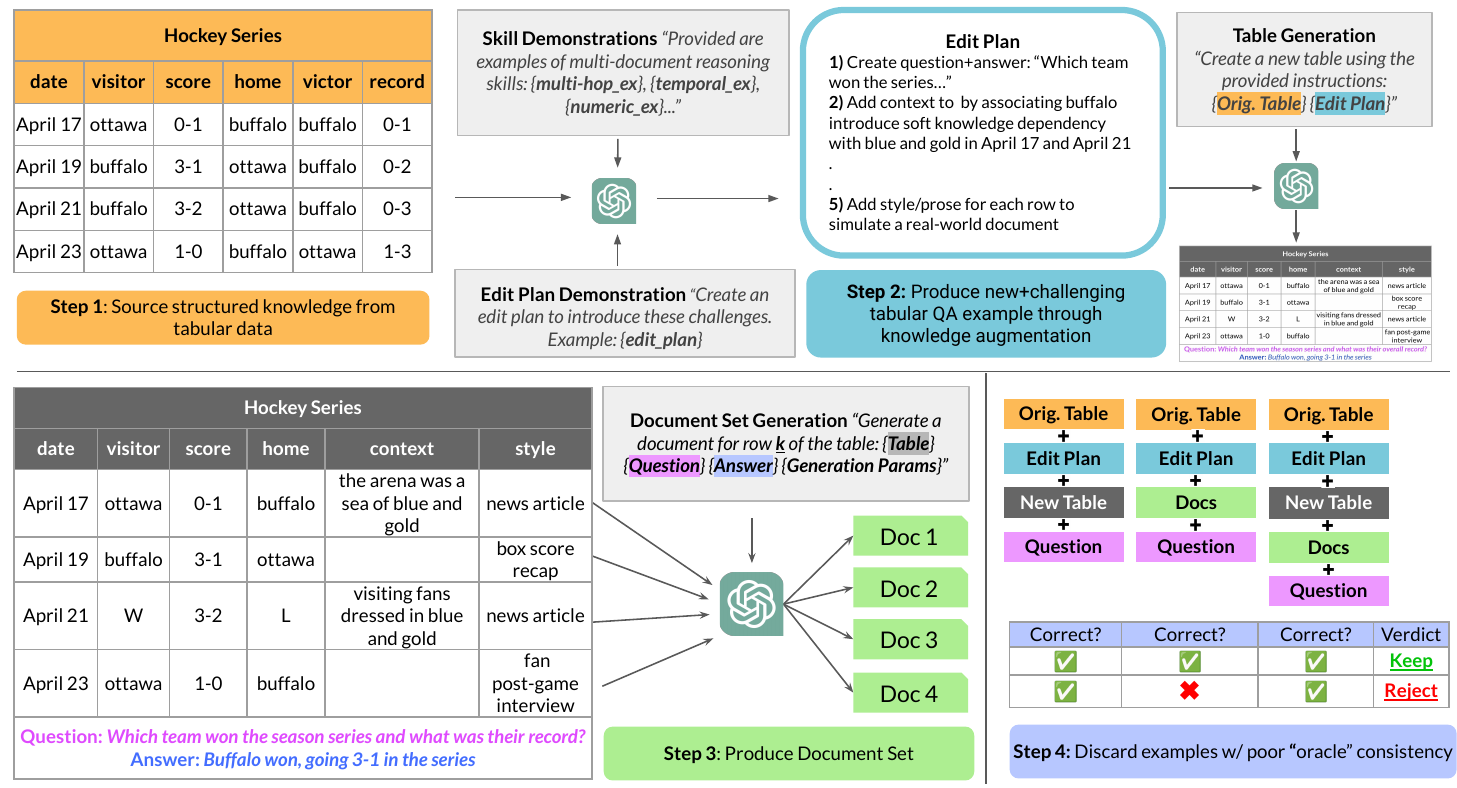}
      \caption{
      \MDName{} generation pipeline overview. We 1) source structured knowledge, then use in-context multi-document reasoning demonstrations to 2) intentionally modify the existing knowledge with challenging dependencies. We then 3) map this seed knowledge into document form to produce the multi-document QA example. Finally, we 4) employ an added validity check to filter inconsistent examples that yield erratic outputs in an oracle setting where example construction knowledge is provided.} 
     \label{fig:pipeline_overview}
\end{figure*}

\section{Introduction}
The rapid advancements in natural language processing (NLP) have been largely driven by the development and deployment of large language models (LLMs). These models have showcased remarkable improvements in various tasks, including understanding, generating, and reasoning over text. However, despite these advancements, evaluation frameworks for NLP systems have struggled to keep pace \citep{chang2024surveyllmevaluation}, notably for tasks involving reasoning over multiple documents~\citep{mavi2024multihopQA}.

Multi-document (MD) reasoning involves synthesizing and inferring information across multiple diverse texts \citep{caciularu2021cdlm}, posing unique challenges not addressed by traditional single-document benchmarks. While LLMs are increasingly capable of handling longer-context multi-document inputs, there is a scarcity of benchmarks that rigorously examine the specific reasoning characteristics that are prominent in this setting. In addition, many existing benchmarks consist of static, hand-crafted datasets, which are labor-intensive to produce. These datasets are often susceptible to data contamination \citep{xu2024benchmarkdatacontaminationlarge} over time, e.g., LLMs are exposed to public benchmarks during training. This can compromise the integrity of the evaluation.

In this work, we address these limitations with \textbf{\MDName{}}\footnote{Benchmark can be found at \url{https://huggingface.co/launch/MDBench}. Code and prompts are accessible at \url{https://github.com/jpeper/MDBench}.}, a benchmark using a novel generation technique for multi-document reasoning evaluation. Our benchmark is generated through a synthetic process that leverages structured knowledge as seed information. This process uses a strong LLM (GPT-4o) to augment structured knowledge by injecting complexities that require advanced reasoning skills, then generates text documents from the augmented knowledge. Synthetic generation allows for efficient example creation and can be easily adapted to incorporate new reasoning skills and objectives.

Our benchmark generation pipeline (overviewed in Figure \ref{fig:pipeline_overview}) begins with a structured knowledge source serving as the seed information. Each knowledge entry (i.e., a row of a table) encapsulates distinct knowledge that forms the basis of a document in the generated set. We use tabular data as these are widely utilized and inherently structured, with row entries often sharing topical connections, making them an ideal foundation for evaluating cross-document reasoning.  We follow a multi-step augmentation process to source knowledge, augment knowledge, and generate high-quality document sets with multi-document reasoning challenges:

\begin{enumerate} \item \textbf{Source Seed Knowledge (Fig. \ref{fig:pipeline_overview}, Step 1)}
We collect tabular data where each row contains information that will contribute to a generated document.
\item \textbf{Augment Knowledge (Fig. \ref{fig:pipeline_overview}, Step 2)} Using GPT-4o, we edit the structured knowledge to inject challenging reasoning dependencies and enrich the context for document creation. By treating rows as proxies for documents, we model cross-document dependencies through cross-row knowledge interactions. In this step, we also generate question-answer pairs that utilize the introduced reasoning dependencies.
\item \textbf{Generate Natural Text (Fig. \ref{fig:pipeline_overview}, Step 3)} We map the augmented knowledge into natural text by generating a corresponding multi-document set from the augmented table. This process allows us to systematically inject critical reasoning challenges while producing examples that are realistic and fluent. 
\item \textbf{Automated Quality Validation (Fig. \ref{fig:pipeline_overview}, Step 4)} To ensure high quality and consistency, we apply targeted consistency prompts and a novel oracle-setting self-consistency check that effectively filters out low-quality examples, enabling rapid benchmark scaling.

\end{enumerate}

We produce 1,000 multi-document QA examples using this pipeline (300 human-validated, and 700 more automatically-validated for quality) and evaluate the performance of models from several prominent LLM families including GPT \cite{openai2024gpt4technicalreport}, Claude \cite{anthropic_claude_3}, Gemini \cite{geminiteam2024geminifamilyhighlycapable}, and LLaMA \cite{dubey2024LLaMA3herdmodels}. We find that:

\begin{itemize} 
\item \MDName{} poses a strong challenge, even for state-of-the-art methods, with the best ones achieving $\sim$81\% accuracy and $\sim$62\% exact-match performance on this MD reasoning task. 

\item Frontier models such as GPT-4o, GPT-o1 and Gemini-2.5-Flash significantly outperform smaller LLMs across different prompting methods. This highlights the importance of model capacity and sophistication in handling complex multi-document reasoning tasks.

\item When comparing performance on document reasoning versus tabular reasoning (i.e., structured format pre-document generation), we find that strong models are mostly performant in both settings, but all struggle more in the document setting. This suggests that \textit{multi-document reasoning is influenced by both the fundamental reasoning complexity, and also from the nuances of the surface form}.

\item Prompting techniques such as chain-of-thought \citep{wei2022chainofthought} can improve performance across strong models. However, they are insufficient to significantly enhance the performance of weaker models like LLaMA-3-7B and even GPT-3.5. This indicates that while prompting strategies can aid reasoning, \textit{underlying model capabilities remain a limiting factor for this task, which makes \MDName{} suitable for future, advanced model evaluation}.
\end{itemize}

\section{Related Work}

Evaluating the capabilities of LLMs is a critical aspect of NLP research. As LLMs continue to improve rapidly, existing evaluation frameworks often lag behind, particularly in assessing complex reasoning abilities such as multi-document (MD) reasoning. As LLMs rapidly increase in reasoning capacity, there is a pressing need to develop evaluation methods that can capture these higher-order reasoning skills. 

\paragraph{Multi-Document Reasoning}
MD reasoning involves synthesizing and inferring information across multiple texts. Existing work in this area includes datasets targeting specific phenomena such as temporal reasoning \citep{xiong2024largelanguagemodelslearntemporal, timedtextranktemporal}, summarization \citep{xiao2021primera, peper2023pelmspretrainingeffectivelowshot, seam_multidocument}, multi-hop question answering with HotPotQA, MuSiQue and 2WikiMultiHop \citep{yang2018hotpotqa, 2wikimultihop, trivedi2022musiquemultihopquestionssinglehop} and ambiguous entity resolution \citep{ambigdocs-lee2024}. Notably, many of these MD datasets are publicly-sourced and often reliant on significant human effort to curate. For example, \citet{zhu2024fanoutqa} introduce FanOutQA, a recent multi-hop, multi-document question answering dataset, which targeted decomposable QA examples sourced from public Wikipedia knowledge and relied on thousands of manual annotations. Our work seeks to use knowledge-controlled generation to offer a scalable alternative for producing nuanced and unseen multi-document reasoning examples.

\paragraph{Tabular Reasoning with LLMs}
LLMs have demonstrated strong performance in tasks involving structured knowledge, such as tabular data or knowledge bases \citep{lu2024large_tabsurvey,li2023sheetcopilot}. Recent studies have observed success in applying LLMs to table reasoning, manipulation, and augmentation \citep{lu2024large_tabsurvey,li2023sheetcopilot}. While there are limitations in LLM pre-training which can lead to formatting sensitivities and limitations with handling large tables, \citet{nahid2024tabsqlify} find improved performance by decomposing the tabular knowledge into a digestible size. Similarly, leveraging tabular knowledge within reasoning chains allows for compact and effective representation of complex problems, as explored in the Chain-of-Tables framework \citep{wang2024chainoftable}. These insights highlight the potential of using condensed knowledge as a foundation for generating challenging reasoning tasks. Although our work draws on structured data to simplify the creation of examples, our ultimate goal is to produce longer, more nuanced natural-language text document sets.

\paragraph{LLM-Supported Synthetic Benchmark Creation}
To address the need for more dynamic evaluation datasets, LLM-powered synthetic benchmark creation has gained significant traction \citep{long2024llmsynthgensurvey, liu2024bestpracticeslessonslearned, li-etal-2023-synthetic}, particularly as there is growing concern of benchmark data contamination \cite{xu2024benchmarkdatacontaminationlarge}. 
Some work has been done in the multi-document setting, although automation is largely used for extending existing annotated multi-document benchmarks to more complex tasks \citep{schnitzler2024morehopqamultihopreasoning}. While not directly modeling multi-document tasks, works like BoardGameQA \cite{kazemi2023boardgameqadatasetnaturallanguage} and MuSR \cite{sprague2023musr} explore synthetic generation in the related multi-step reasoning setting, with MuSR using a neurosymbolic generation algorithm which maps synthetic structure into natural text examples. Our method seeks to build off related work in synthetic generation to address efficient multi-document benchmark creation.

\section{\MDName{} Generation Pipeline}

In this section, we motivate and overview the generation process, and provide details on the components and steps taken to produce the \MDName{} evaluation benchmark.

Our design decisions for the \MDName{} benchmark are guided by several key objectives to ensure the resulting examples test a broad and relevant range of reasoning skills:

\begin{itemize}
    \item \textbf{Contain Novel and Unseen Text}: We aim to produce examples that are not merely scraped from public datasets but rather contain newly-generated content. This ensures that models are tested on scenarios they have not encountered during training, avoiding overfitting to pre-existing benchmarks.
    \item \textbf{Contains Cross-Document Knowledge Dependencies}: A key focus is to produce examples that require reasoning across multiple documents. We design our benchmark to have intentional cross-document dependencies, making them particularly challenging to effectively test multi-document reasoning capabilities. 
    \item \textbf{Grounded in Real-World Scenarios}: Even though the examples are synthetically generated, they should ideally remain grounded in real-world concepts and situations. This ensures that the reasoning challenges presented are realistic and relevant to practical NLP applications.
    \item \textbf{Counterfactual Alterations}: To further mitigate data contamination and leakage risks from public sources, we allow slight counterfactual or fictional twists on the real-world scenarios occurring in the seed dataset to necessitate that models leverage the extrinsic knowledge found in the input. This allows for a fresh take on familiar domains while maintaining the integrity of the benchmark.
    \item \textbf{Scalability and Control}: Our approach is designed to offer control during benchmark generation. We allow one to specify seed information such as domain and behavior types, and can control the complexity and nature of the reasoning tasks present in the benchmark.
\end{itemize}

\begin{table}[]
    \small
    \centering
    \begin{tabular}{l|rcr}
\hline
                              & Min  & Mean (std)      & \multicolumn{1}{l}{Max} \\ \hline
\# Rows per Table                & 5    & 8.31 (2.4)      & 17                      \\
\# Table Columns              & 3    & 5.17 (1.2)      & 9                       \\
Token Length (Tab.) & 121  & 256.0 (81.6)   & 554                     \\
Token Length (Doc.)    & 1048 & 2397.4 (738.4) & 6493                    \\
Avg. Doc Length    & 177  & 268.1 (37.6)   & 388                     \\ \hline
\end{tabular}
    \caption{
    \MDName{} benchmark statistics. Each row in the tabular representation ultimately corresponds to a document within the multi-document example. We see a roughly nine-fold increase in surface form length when mapping the structured knowledge to natural text document format.}
    \label{tab:dataset_stats}
\end{table}

\begin{table*}[t]
    \small
    \centering
    \begin{tabular}{|>{\centering\arraybackslash}m{4cm}|>{\centering\arraybackslash}m{8cm}|}
\hline
\textbf{Reasoning Type} & \textbf{Description} \\ \hline
Multi-hop Reasoning & Solving problems requiring multiple steps to arrive at the solution. \\ \hline
Numeric Reasoning & Handling numeric values and performing numerical operations. \\ \hline
Temporal Reasoning & Handle temporal information and temporal dependencies. \\ \hline
Knowledge Aggregation & Aligning, comparing and/or contrasting knowledge that may be present. \\ \hline
Soft Reasoning & Reasoning abductively and making informed decisions in cases where some uncertainty or fuzziness may be present, such as cross-document entity linking. \\ \hline
\end{tabular}
    \caption{
    Reasoning skills overview. For our benchmark, we focus on five goals which are especially relevant for the multi-document setting. We provide demonstrations of these reasoning types to inspire relevant knowledge edits during the generation process.}
    \label{tab:reasoning_types}
\end{table*}
\begin{figure}[h]
\centering
     \includegraphics[width=\columnwidth]{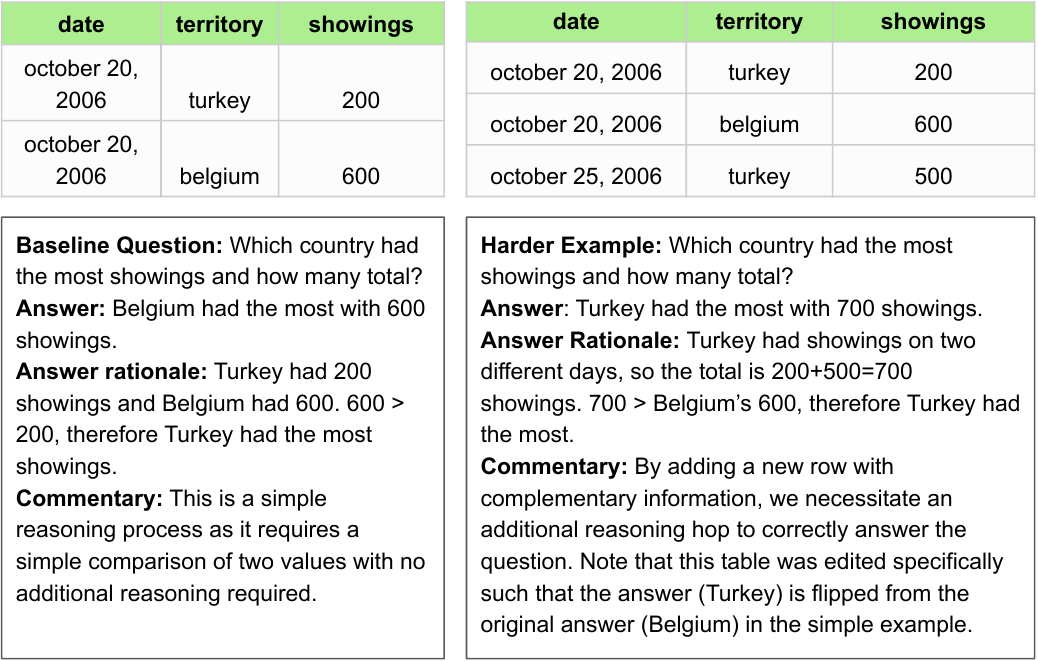}
      \caption{Example skill description on ``Multi-hop Reasoning''. During knowledge augmentations, we demonstrate the multi-document skills relevant to the document sets.}
     \label{fig:skill_description_formatted}
\end{figure}

\subsection{Pipeline Overview}
Our benchmark generation pipeline begins with structured knowledge sourced from tabular data, which serves as the seed for the augmentation process. This structured knowledge is systematically enriched and refined through a strong LLM to inject reasoning dependencies that challenge models to infer information across multiple documents. Figure \ref{fig:pipeline_overview} overviews the pipeline. Below, we provide details of the four major steps:

\paragraph{Step 1: Obtaining Seed Knowledge}

We start with an intuition that compressed structured knowledge provides an effective foundation for multi-document reasoning. Several valid sources of this exist, such as knowledge bases, tabular information, or even by performing information extraction to consolidate data from existing documents and text corpora. For the \MDName{} benchmark, we utilize the TabFact \citep{chen2020tabfactlargescaledatasettablebased} dataset, which comprises 16,000 tables sourced from Wikipedia. Our motivation for exploring this dataset is threefold: (1) TabFact tables provide a reliable and curated source of seed knowledge; (2) the data spans a wide range of domains, including news, sports, media, and technology; and (3) it has an emphasis on human-readability both in scale and content. This structured knowledge serves as the starting point for our knowledge augmentation process, which significantly transforms the raw data into more challenging and complex reasoning tasks. We heuristically filter the dataset to select tables which are rich in content yet manageable in size, choosing those with 5 to 17 rows and 3 to 9 columns.

\paragraph{Step 2: Knowledge Augmentation}
An important component of our technique is the knowledge augmentation step. This step modifies information, applying operations that inject complex knowledge dependencies and reasoning challenges. Figure \ref{fig:pipeline_overview} overviews our pipeline, while full detailed examples of the knowledge augmentation prompts are provided in Appendix \ref{sec:appendix_reasoning_demonstrations}. Below, we describe two key techniques we employ:

\begin{itemize}
\item \textbf{Multi-document Reasoning Demonstrations} Prior to altering the existing information, we first demonstrate relevant skills for multi-document reasoning. Each skill is demonstrated in both `simple' and `challenging' forms (e.g., Figure \ref{fig:skill_description_formatted}). The demonstrations include examples, along with explanations and rationales for solving them.
For the purpose of this benchmark, we define and focus on five reasoning components which are particularly relevant in the multi-document setting: (1) \texttt{Multi-hop}, (2) \texttt{Numeric}, (3) \texttt{Temporal}, (4) \texttt{Soft reasoning}, and (5) \texttt{Knowledge Aggregation}.

We briefly describe these skills in Table~\ref{tab:reasoning_types}, and provide full demonstrations of each in Appendix \ref{sec:appendix_reasoning_demonstrations}.

\item \textbf{Knowledge Augmentation Demonstrations} In addition to demonstrating relevant reasoning skills, we next provide \textit{knowledge edit demonstrations}. These demonstrations illustrate plans for how simple tables can be enhanced to form nuanced QA examples. Each demonstration consists of an initial table, a series of edits, and a resultant augmented table and QA annotation. When performing knowledge augmentation, we sample one demonstration from a small set of high-quality curated examples. Figure \ref{fig:edit_plan_ex} demonstrates a knowledge edit plan.
\end{itemize}

Through these two steps, we modify the tabular knowledge to form a more nuanced QA example with cross-row knowledge dependencies.

\paragraph{Step 3: Document Set Generation}
Once the tabular knowledge has been augmented, we map this information into natural language text; each row in the table is used to generate a document, with the augmented knowledge ensuring that reasoning across documents (rows) is required to solve the accompanying QA task. We independently generate each document, with the generation prompt parameterized by the following components: (1) the augmented table and title, (2) the column names, and (3) a specific row of content within the table indicated for generation. Iterating this process over all $n$ rows in the table, we generate an $n$-document set. This approach of knowledge-grounded generation ensures the generated document set maintains logical coherence while presenting unique cross-document reasoning challenges. Figures \ref{fig:example_1} and \ref{fig:example_2} demonstrate a full generated \MDName{} multi-document example.

\paragraph{Step 4: Automated Quality Validation}

To ensure a high-quality and scalable benchmark, our pipeline employs a multi-step automated validation process. First, we apply targeted prompts (Appendix ~\ref{sec:appendix_validity_prompts}) to verify that each component—the seed table, edit plans, intermediate edited table, and final document set—adheres to the intended instructions. Next, we perform an oracle self-consistency check based on the intuition that the procedural steps used to construct an example can also verify its consistency. Specifically, for each QA example, we prompt GPT-4o under three variations of \textit{oracle knowledge} (original table and edit plan) and \textit{context} (generated table and generated document set) as shown in Figure \ref{fig:pipeline_overview}.
An example is rejected if \textit{any} of the three variations yield an answer which does not match the generated ground-truth. This comprehensive validation ensures internal consistency and \textit{underpins scalable, automated quality control}.

\subsection{\MDName{} Benchmark Generation Details}
We utilize GPT-4o for both table augmentation and document generation. Table~\ref{tab:dataset_stats} summarizes the statistics of the resultant benchmark. We produce a benchmark comprising 1,000 examples.

\paragraph{Automated Validation}
Our pipeline integrates automated validation (including both localized validation checks as well as the oracle self-consistency check) to ensure quality and consistency throughout generation. During generation, we find the automated filtering retained approximately 32\% of the generated examples.

\paragraph{Human Validation}
To assess the quality of the generated and auto-verified examples, we hire and train a native English-speaking graduate student with linguistics background for this task (compensated at \$15 per hour for training and annotation). The annotator is instructed to validate the consistency of the generated multi-document QA examples (see Appendix \ref{fig:annotation_guidelines}). We find our automated validation yields an overall example validity of rate of 87\%. Our final released benchmark comprises 300 human-validated examples and 700 additional machine-validated examples.

\begin{table*}[h]
    \centering
    \begin{tabular}{
>{\columncolor[HTML]{FCFCFC}}l |ccccc}
\hline
\textbf{Model}       & \cellcolor[HTML]{FCFCFC}\textbf{Zero-shot} & \cellcolor[HTML]{FCFCFC}\textbf{Zero-shot CoT} & \cellcolor[HTML]{FCFCFC}\textbf{One-shot} & \cellcolor[HTML]{FCFCFC}\textbf{One-shot CoT} & \cellcolor[HTML]{FCFCFC}\textbf{Overall} \\ \hline
LLaMA-3-8B-Instruct  & \cellcolor[HTML]{F0BABA}42.7               & \cellcolor[HTML]{F1BCBC}43.1                   & \cellcolor[HTML]{EDA8A8}39.3              & \cellcolor[HTML]{ECA5A5}38.9                  & \cellcolor[HTML]{EEB1B1}41.0             \\
LLaMA-3-70B-Instruct & \cellcolor[HTML]{FAE8E8}51.2               & \cellcolor[HTML]{F9E2E2}50.2                   & \cellcolor[HTML]{F3C6C6}45.0              & \cellcolor[HTML]{F8DDDD}49.3                  & \cellcolor[HTML]{F7DBDB}48.9             \\
Claude-3-Opus        & \cellcolor[HTML]{FFFFFF}55.5               & \cellcolor[HTML]{FDF9F9}54.5                   & \cellcolor[HTML]{E7F6EF}56.4              & \cellcolor[HTML]{FCF4F4}53.6                  & \cellcolor[HTML]{FEFCFC}55.0             \\
Claude-3.5-Sonnet    & \cellcolor[HTML]{9FD9BD}59.2               & \cellcolor[HTML]{E7F6EF}56.4                   & \cellcolor[HTML]{B8E2CD}58.3              & \cellcolor[HTML]{F3FBF7}55.9                  & \cellcolor[HTML]{CCEBDC}57.5             \\
Gemini-1.5-Pro       & \cellcolor[HTML]{FDF7F7}54.0               & \cellcolor[HTML]{C4E7D6}57.8                   & \cellcolor[HTML]{FDF7F7}54.0              & \cellcolor[HTML]{FDF7F7}54.0                  & \cellcolor[HTML]{FEFCFC}55.0             \\
Gemini-2.5-Flash     & \cellcolor[HTML]{C4E7D6}57.8               & \cellcolor[HTML]{B7E2CD}58.3                   & \cellcolor[HTML]{ABDDC4}58.8              & \cellcolor[HTML]{87CFAC}\textbf{60.2}         & \cellcolor[HTML]{ABDDC4}58.8             \\
GPT-3.5-Turbo        & \cellcolor[HTML]{F8E0E0}49.8               & \cellcolor[HTML]{EB9E9E}37.4                   & \cellcolor[HTML]{F2C4C4}44.5              & \cellcolor[HTML]{EA9999}36.5                  & \cellcolor[HTML]{F0B6B6}42.1             \\
GPT-4o               & \cellcolor[HTML]{93D4B4}\textbf{59.7}      & \cellcolor[HTML]{57BB8A}\textbf{62.1}          & \cellcolor[HTML]{93D4B4}\textbf{59.7}     & \cellcolor[HTML]{B8E2CD}58.3                  & \cellcolor[HTML]{8ED1B0}\textbf{60.0}    \\
GPT-o1               & \cellcolor[HTML]{D1EDDF}57.3               & \cellcolor[HTML]{7BCAA3}60.7                   & \cellcolor[HTML]{94D4B5}\textbf{59.7}     & \cellcolor[HTML]{A1D9BD}59.2                  & \cellcolor[HTML]{A1D9BD}59.2             \\ \hline
\end{tabular}

    \begin{tabular}{
>{\columncolor[HTML]{FCFCFC}}l |ccccc}
\hline
\textbf{Model}       & \cellcolor[HTML]{FCFCFC}\textbf{Zero-shot} & \cellcolor[HTML]{FCFCFC}\textbf{Zero-shot CoT} & \cellcolor[HTML]{FCFCFC}\textbf{One-shot} & \cellcolor[HTML]{FCFCFC}\textbf{One-shot CoT} & \cellcolor[HTML]{FCFCFC}\textbf{Overall} \\ \hline
LLaMA-3-8B-Instruct  & \cellcolor[HTML]{F2C3C3}65.7               & \cellcolor[HTML]{F3C9C9}66.8                   & \cellcolor[HTML]{EDAAAA}60.7              & \cellcolor[HTML]{F0B8B8}63.5                  & \cellcolor[HTML]{F1BBBB}64.2             \\
LLaMA-3-70B-Instruct & \cellcolor[HTML]{FDF7F7}76.1               & \cellcolor[HTML]{FDF5F5}75.6                   & \cellcolor[HTML]{F4CBCB}67.4              & \cellcolor[HTML]{F8E0E0}71.6                  & \cellcolor[HTML]{F9E6E6}72.7             \\
Claude-3-Opus        & \cellcolor[HTML]{CDEBDC}79.3      & \cellcolor[HTML]{FBEEEE}74.2                   & \cellcolor[HTML]{DAF0E6}78.8              & \cellcolor[HTML]{FBEEEE}74.2                  & \cellcolor[HTML]{FEFAFA}76.6             \\
Claude-3.5-Sonnet    & \cellcolor[HTML]{F2FAF6}78.0               & \cellcolor[HTML]{FEFEFE}77.3                   & \cellcolor[HTML]{AEDFC7}80.3              & \cellcolor[HTML]{F6FCF9}77.9                  & \cellcolor[HTML]{E7F5EE}78.4             \\
Gemini-1.5-Pro       & \cellcolor[HTML]{FEFCFC}77.1               & \cellcolor[HTML]{B9E3CF}80.0                   & \cellcolor[HTML]{FCFEFD}77.7              & \cellcolor[HTML]{FCF1F1}74.9                  & \cellcolor[HTML]{FEFEFE}77.4             \\
Gemini-2.5-Flash     & \cellcolor[HTML]{F5FBF8}77.9               & \cellcolor[HTML]{CFECDE}79.2                   & \cellcolor[HTML]{C1E6D4}79.7              & \cellcolor[HTML]{C1E6D4}79.7                  & \cellcolor[HTML]{D2EDE0}79.1             \\
GPT-3.5-Turbo        & \cellcolor[HTML]{F7DADA}70.4               & \cellcolor[HTML]{EA9999}57.3                   & \cellcolor[HTML]{F3C6C6}66.3              & \cellcolor[HTML]{EBA0A0}58.7                  & \cellcolor[HTML]{F0B6B6}63.2             \\
GPT-4o               & \cellcolor[HTML]{D2EDE0}79.1               & \cellcolor[HTML]{BDE5D1}79.8                   & \cellcolor[HTML]{A2DABE}80.8              & \cellcolor[HTML]{E8F6EF}78.3                  & \cellcolor[HTML]{C6E8D8}79.5             \\
GPT-o1               & \cellcolor[HTML]{98D5B7}\textbf{81.1}      & \cellcolor[HTML]{57BB8A}\textbf{83.3}          & \cellcolor[HTML]{78C8A1}\textbf{82.2}     & \cellcolor[HTML]{72C69D}\textbf{82.4}         & \cellcolor[HTML]{78C8A1}\textbf{82.2}    \\ \hline
\end{tabular}

    \caption{
    Document Reasoning Overall Results. We report exact-match (top) and accuracy (bottom) results on the \MDName{} multi-document examples.
    }
    \label{tab:doc_gen_main_results}
\end{table*}

\begin{table*}[t]
    \centering
    \begin{tabular}{
>{\columncolor[HTML]{FCFCFC}}l |ccccc}
\hline
\textbf{Model}       & \cellcolor[HTML]{FCFCFC}\textbf{Zero-shot} & \cellcolor[HTML]{FCFCFC}\textbf{Zero-shot CoT} & \cellcolor[HTML]{FCFCFC}\textbf{One-shot} & \cellcolor[HTML]{FCFCFC}\textbf{One-shot CoT} & \cellcolor[HTML]{FCFCFC}\textbf{Overall} \\ \hline
LLaMA-3-8B-Instruct  & \cellcolor[HTML]{F2C3C3}38.4               & \cellcolor[HTML]{F5D3D3}45.0                   & \cellcolor[HTML]{EA9999}19.9              & \cellcolor[HTML]{F3C7C7}39.8                  & \cellcolor[HTML]{F1BDBD}35.8             \\
LLaMA-3-70B-Instruct & \cellcolor[HTML]{FBEFEF}57.3               & \cellcolor[HTML]{EEF8F3}64.9                   & \cellcolor[HTML]{EFB2B2}30.8              & \cellcolor[HTML]{FBEDED}56.4                  & \cellcolor[HTML]{F9E4E4}52.4             \\
Claude-3-Opus        & \cellcolor[HTML]{E5F5ED}65.4               & \cellcolor[HTML]{FEFAFA}62.1                   & \cellcolor[HTML]{DCF1E7}65.9              & \cellcolor[HTML]{D3EEE1}66.4                  & \cellcolor[HTML]{EEF8F3}64.9             \\
Claude-3.5-Sonnet    & \cellcolor[HTML]{9ED8BC}69.2               & \cellcolor[HTML]{C2E6D4}67.3                   & \cellcolor[HTML]{C2E6D4}67.3              & \cellcolor[HTML]{FFFFFF}64.0                  & \cellcolor[HTML]{C8E9D9}66.9             \\
Gemini-1.5-Pro       & \cellcolor[HTML]{FEFCFC}63.0               & \cellcolor[HTML]{C2E6D4}67.3                   & \cellcolor[HTML]{FEFAFA}62.1              & \cellcolor[HTML]{FFFFFF}64.0                  & \cellcolor[HTML]{FDFFFE}64.1             \\
Gemini-2.5-Flash     & \cellcolor[HTML]{FFFFFF}64.0                   & \cellcolor[HTML]{E5F5ED}65.4                   & \cellcolor[HTML]{FEFDFD}63.5              & \cellcolor[HTML]{EEF9F4}64.9                  & \cellcolor[HTML]{F4FBF7}64.5             \\
GPT-3.5-Turbo        & \cellcolor[HTML]{F8E1E1}51.2               & \cellcolor[HTML]{F8DEDE}49.8                   & \cellcolor[HTML]{FAE9E9}54.5              & \cellcolor[HTML]{F9E4E4}52.6                  & \cellcolor[HTML]{F9E3E3}52.0             \\
GPT-4o               & \cellcolor[HTML]{72C69D}\textbf{71.6}      & \cellcolor[HTML]{9ED8BC}\textbf{69.2}          & \cellcolor[HTML]{57BB8A}\textbf{73.0}     & \cellcolor[HTML]{7BCAA3}\textbf{71.1}         & \cellcolor[HTML]{79C9A2}\textbf{71.2}    \\
GPT-o1               & \cellcolor[HTML]{D4EEE1}66.3               & \cellcolor[HTML]{A7DCC2}68.7                   & \cellcolor[HTML]{95D4B5}69.7              & \cellcolor[HTML]{CBEADB}66.8                  & \cellcolor[HTML]{B6E2CD}67.9             \\ \hline
\end{tabular}

    \begin{tabular}{
>{\columncolor[HTML]{FCFCFC}}l |ccccc}
\hline
\textbf{Model}       & \cellcolor[HTML]{FCFCFC}\textbf{Zero-shot} & \cellcolor[HTML]{FCFCFC}\textbf{Zero-shot CoT} & \cellcolor[HTML]{FCFCFC}\textbf{One-shot} & \cellcolor[HTML]{FCFCFC}\textbf{One-shot CoT} & \cellcolor[HTML]{FCFCFC}\textbf{Overall} \\ \hline
LLaMA-3-8B-Instruct  & \cellcolor[HTML]{F6D5D5}59.0               & \cellcolor[HTML]{F8E1E1}65.1                   & \cellcolor[HTML]{EA9999}28.0              & \cellcolor[HTML]{F4CCCC}54.1                  & \cellcolor[HTML]{F3C7C7}51.5             \\
LLaMA-3-70B-Instruct & \cellcolor[HTML]{FDF5F5}75.3               & \cellcolor[HTML]{FEFDFD}79.4                   & \cellcolor[HTML]{EEAFAF}39.7              & \cellcolor[HTML]{FCF2F2}73.5                  & \cellcolor[HTML]{F9E5E5}66.9             \\
Claude-3-Opus        & \cellcolor[HTML]{FEFEFE}79.8               & \cellcolor[HTML]{F5FBF8}80.4                   & \cellcolor[HTML]{FCFEFD}80.1              & \cellcolor[HTML]{F6FCF9}80.3                  & \cellcolor[HTML]{FAFDFC}80.1             \\
Claude-3.5-Sonnet    & \cellcolor[HTML]{9BD7B9}84.1               & \cellcolor[HTML]{AEDFC7}83.3                   & \cellcolor[HTML]{BBE4D0}82.8              & \cellcolor[HTML]{FEFEFE}79.7                  & \cellcolor[HTML]{C2E7D5}82.5             \\
Gemini-2.5-Flash     & \cellcolor[HTML]{FCFCFC}79.3               & \cellcolor[HTML]{FEFCFC}78.8                   & \cellcolor[HTML]{FEFDFD}79.4              & \cellcolor[HTML]{FEFEFE}79.6                  & \cellcolor[HTML]{FEFDFD}79.3             \\
Gemini-1.5-Pro       & \cellcolor[HTML]{ECF7F2}80.8               & \cellcolor[HTML]{D1EDDF}81.8                   & \cellcolor[HTML]{FEFBFB}78.3              & \cellcolor[HTML]{E5F5ED}81.0                  & \cellcolor[HTML]{F2FAF6}80.5             \\
GPT-3.5-Turbo        & \cellcolor[HTML]{FAE9E9}69.1               & \cellcolor[HTML]{FAE7E7}67.9                   & \cellcolor[HTML]{FBEBEB}70.0              & \cellcolor[HTML]{F9E6E6}67.3                  & \cellcolor[HTML]{FAE8E8}68.6             \\
GPT-4o               & \cellcolor[HTML]{74C79E}\textbf{85.7}      & \cellcolor[HTML]{D3EDE0}81.8                   & \cellcolor[HTML]{57BB8A}\textbf{86.9}     & \cellcolor[HTML]{A8DCC3}83.6                  & \cellcolor[HTML]{92D3B3}\textbf{84.5}    \\
GPT-o1               & \cellcolor[HTML]{92D3B3}84.5               & \cellcolor[HTML]{88CFAC}\textbf{84.9}          & \cellcolor[HTML]{AADDC4}83.5              & \cellcolor[HTML]{9BD7BA}\textbf{84.1}         & \cellcolor[HTML]{99D6B8}84.2             \\ \hline
\end{tabular}

    \caption{Table Reasoning Overall Results. We report exact-match (top) and accuracy (bottom) when applying models to the augmented \textit{tabular format} QA examples (as opposed to documents).
    }
    \label{tab:table_gen_main_results}
\end{table*}

\section{Experimental Setup}
To assess the challenges of \MDName{}, we test the performance of many popular LLMs in combination with conventional prompting setups.
Concretely, we test open-weights LLMs with Meta's \textbf{LLaMA-3} \citep{dubey2024LLaMA3herdmodels}, using the 8B-Instruct and 70B-Instruct variants.  \
For API-based proprietary methods, we use models from the Anthropic \textbf{Claude}, OpenAI \textbf{GPT}, and Google \textbf{Gemini} families, which represent current state-of-the-art LLMs.
For Claude we evaluate Claude-3-Opus \texttt{20240229} and Claude-3.5-Sonnet \texttt{20240620}\footnote{\url{https://www.anthropic.com/claude}}.
For GPT we test GPT-3.5-Turbo-16k \texttt{0613}\footnote{\url{https://platform.openai.com/docs/models/gpt-3-5}} \citep{ouyang2022traininglanguagemodelsfollow}, GPT-4o \texttt{2024-08-06}\footnote{\url{https://platform.openai.com/docs/models/gpt-4o}}, and GPT-o1 \texttt{2024-12-17}\footnote{\url{https://platform.openai.com/docs/models/o1}}.
For Gemini we include Gemini-1.5-Pro \texttt{0514} \citep{geminiteam2024gemini15unlockingmultimodal} and the recent Gemini-2.5-Flash-preview \texttt{05-20}\footnote{\url{https://ai.google.dev/gemini-api/docs/models/2.5-flash}}.

\begin{figure}[]
\centering
     \includegraphics[width=\columnwidth]{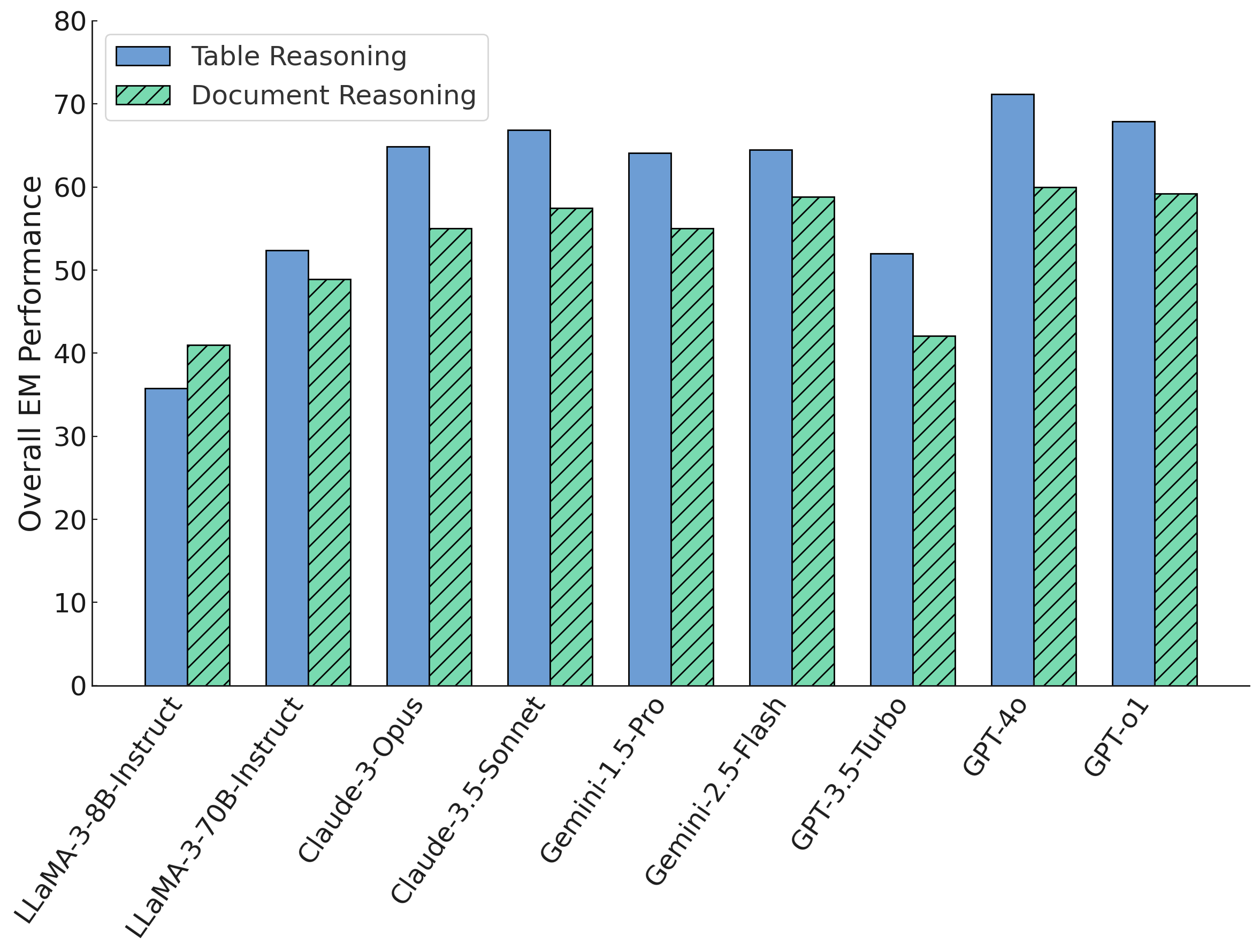}
      \caption{Overall exact-match performance of models on \MDName{}. Table Reasoning is when evaluated with the intermediate table QA examples. Document Reasoning refers to the performance on the \textit{final task} of multi-document reasoning. Document reasoning poses a harder challenge, speaking to the importance of surface form in reasoning performance.}
     \label{fig:overall_performance}
\end{figure}

\begin{figure*}[]
\centering
     \includegraphics[width=\textwidth]{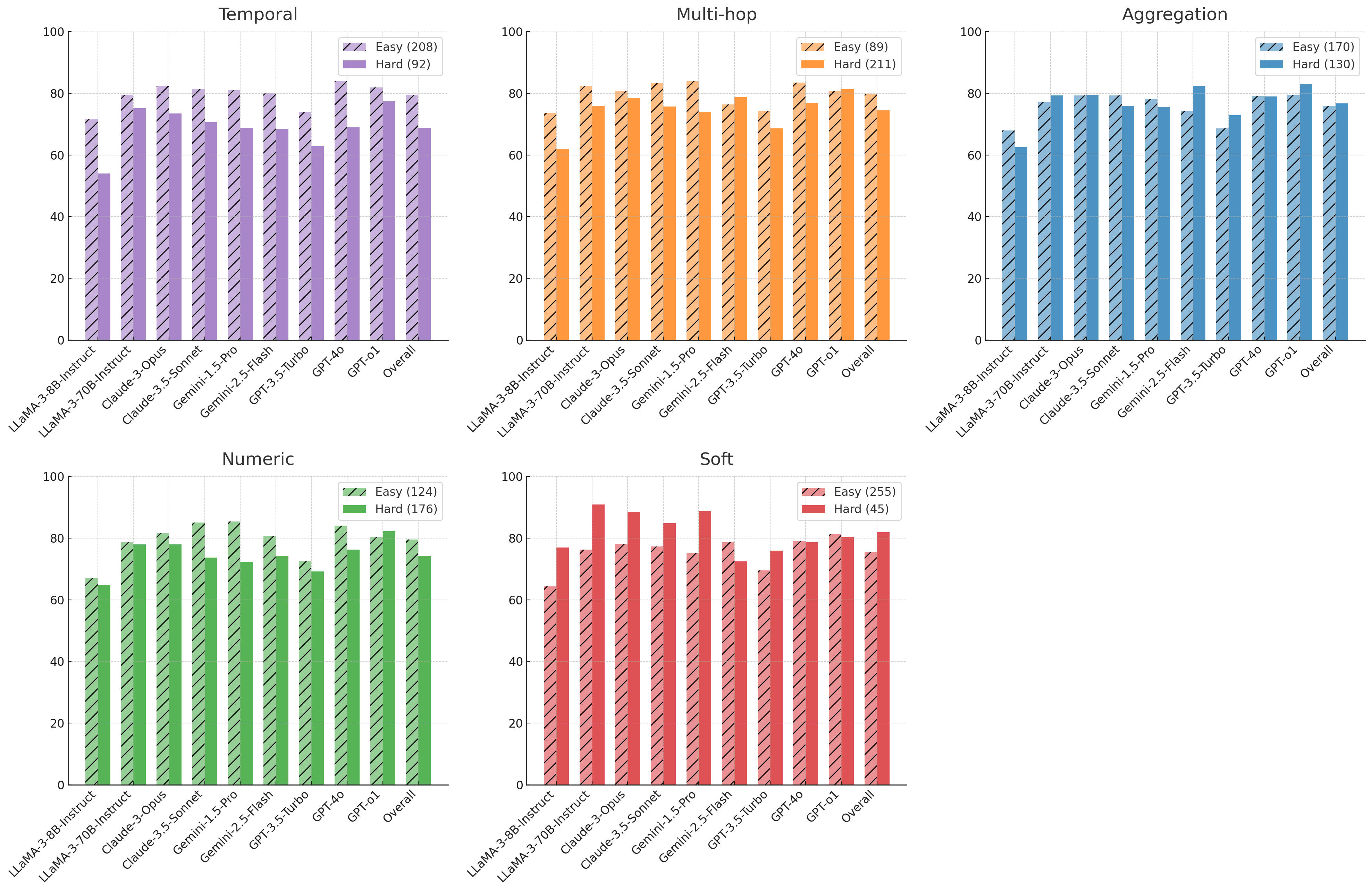}
      \caption{
      Characteristic-level performance breakdown. We partition the examples by difficulty for each characteristic (as assessed by an LLM judge) and report each model's overall accuracy on each of the bins (simple and hard).}
     \label{fig:characteristic_results}
\end{figure*}

\begin{table}[]
  \centering

  \begin{tabular}{l|cc}
    \hline
    \textbf{Ablation} &
     \textbf{GPT-4o} &
     \textbf{GPT-3.5} \\ 
     \hline
    Original Doc. Set               & 59.7 & 49.8 \\
    - No Delimiter     & 55.5 & 45.0 \\
    - Shuffled               & 53.1 & 39.3 \\
    - No Delimiter + Shuff.     & 50.2 & 41.7 \\ 
    \hline
    \end{tabular}
    \caption{We analyze the significance of document ordering and delimitation within our benchmark (in the zero-shot setting). We see that both shuffling the order of the original docs, and removing ``Document $\langle num \rangle$'' delimiters impact performance, implying the existence of temporal and cross-document dependencies within our reasoning problems.}
    \label{tab:ordering_sensitivity}
\end{table}

We explore both \textit{zero-shot} and \textit{one-shot} QA prompting scenarios, noting that when prompting in the one-shot case we use a single representative demonstration across models for consistency. We use a conventional question-answering prompt, and also further instruct the models to `think step by step' to additionally produce \textit{chain-of-thought} (CoT) rationales. Examples of these prompt formats are provided in Appendix \ref{sec:appendix_llm_prompts}. To evaluate on the QA task, we use GPT-4o as a reference-based scorer, first parsing the final answer from each output, then comparing the similarity of the predicted answer with the ground-truth answer (conditioned on the original question). We calculate both an \textit{exact match} score as well as an \textit{accuracy} score, where the scorer can assign partial correctness credit (0-10 scale) for multi-part questions. Full post-processing and scoring prompts are provided in
Appendix \ref{sec:appendix_postproc_prompts}.

\section{Results and Analysis}

\paragraph{Overall Findings}
Figure~\ref{fig:overall_performance} overviews the overall task performance on the multi-document and intermediate tabular versions of the dataset, while Table~\ref{tab:doc_gen_main_results} covers the detailed performance in the target multi-document setting. \MDName{} poses a strong challenge, with the best methods achieving roughly 60\% exact-match performance. \textit{GPT-4o leads with 60.0\% overall EM} and 62.1\% in the zero-shot CoT setting, with GPT-o1 close behind at 59.2\% EM. Notably, we see \textit{o1 attains the highest accuracy score across the board in the document setting} (82.2\% overall). We generally observe that o1, with its deep reasoning capacities, performs better in the CoT and one-shot settings than in zero-shot.
Gemini-1.5 Pro struggles relatively amongst proprietary models, however, the recent Gemini-2.5-Flash improves consistently over it in most settings, peaking at 60.2\% EM under one-shot CoT. Overall, we see relatively strong performance from one or more models from each family, with Claude-3.5-Sonnet performing best among the Claude family. The use of CoT prompting yields performance gains in larger models like Gemini-Flash-2.5, GPT-o1 and GPT-4o, although the effect varies. We see open-weight models struggle overall, although LLaMA-3-70B bests GPT-3.5-Turbo, with the 8B variant comparable in performance to GPT-3.5.

\paragraph{Document vs. Tabular Reasoning}
To ascertain the impact of surface form on the reasoning task, we compare the performance of models on the full multi-document version of the benchmark versus the table version (i.e., stopping after step 2 in our pipeline). Table \ref{tab:table_gen_main_results} overviews the table-reasoning results, and the comparison of overall results can be seen in Figure \ref{fig:overall_performance}. We observe that performance is noticeably higher on the condensed tabular format than in the document setting, with GPT-4o achieving the highest table-reasoning performance at 71.2\% overall exact-match, and GPT-o1 similarly performing strongly in the accuracy metric. A key exception to the tabular vs. document trend is with the LLaMA models, which struggle considerably more on the tabular reasoning task, performing significantly worse than GPT-3.5 in this setting.

\paragraph{Characteristic Breakdown}
We additionally evaluate the performance as a function of the example difficulty. To do this, we prompt GPT-4o to generate characteristic-level difficulty scores for each example. We use the same five characteristics as demonstrated in the generation process, and prompt the model with these definitions. Rather than generating absolute scores, we instead approximate difficulty by prompting GPT-4o to perform comparative ranking with two other randomly sampled examples for each characteristic. We aggregate these relative rankings over the entire dataset to form two difficulty bins per characteristic, as overviewed in Figure \ref{fig:characteristic_results}. 

We see mostly consistent trends across characteristics, with temporal reasoning posing the steepest dropoff between the simple and hard bins. Interestingly, we see soft reasoning is impacted inversely, with performance increasing on the split of examples ranked to have harder soft-reasoning components. While some of this may be due to small sample size for for the hard bin (only 45 of 300 examples), we suspect there is an inverse relationship between soft reasoning and more `explicit' characteristics such as numeric and temporal. For example, a table/example well-suited for temporal reasoning may naturally contain less `soft' information requirements. Conversely, an example with significant soft reasoning requirements likely contains fewer hard reasoning requirements.

\vspace{30pt} 

\paragraph{Document Ordering Analysis}
Finally, we conduct a brief case study with GPT-4o and GPT-3.5 to assess the significance of document ordering.
Table \ref{tab:ordering_sensitivity} shows that both explicit document delimiters and the canonical ordering of the documents are substantive cues for current LLMs. Removing the “Document $\langle num \rangle$” tags lowers GPT-4o accuracy by 4.2 points and GPT-3.5 by 4.8 points, indicating that models leverage these textual anchors to partition the multi-document contexts reliably. Shuffling the document order (while keeping delimiters) produces an even steeper decline (6.6 points for GPT-4o, 10.5 points for GPT-3.5), suggesting that the questions often rely on temporal or positional relationships established by the original document sequence. When both cues are ablated simultaneously, performance drops even further relative to the baseline. Taken together, the results confirm that \MDName{} examples are not solvable by treating each document in isolation. Rather, models must track cross-document dependencies that are sensitive to both ordering and explicit boundary markers.

\section{Conclusion}
We present \MDName{}, a novel benchmark for evaluating multi-document reasoning in LLMs. Leveraging structured seed knowledge and augmenting it with nuanced reasoning dependencies, \MDName{} enables the systematic creation of challenging QA examples while addressing data contamination and annotation costs. Our results reveal significant challenges for current models, and our approach facilitates scalable, targeted evaluation of multi-document reasoning capabilities, paving the way for more rigorous evaluation of models’ abilities to handle real-world, multi-source information.

\section*{Acknowledgments}
This work is supported by Cisco Research. We thank Greg Durrett, Grace Kim, Ramya Namuduri, Liyan Tang and Anisha Gunjal for their insights and valuable discussions.  We also thank the reviewers of ICLR and ARR for their thoughtful comments. 

\section*{Limitations}

While \MDName{} presents a novel approach to evaluating multi-document reasoning in large language models, some directions warrant further exploration: First, we evaluate this method in an automation-focused setting, manually defining just a few foundational reasoning skills and demonstrations as prompts. Future work could explore the trade-offs in adopting more human-centric benchmark processes, such as identifying which pipeline components yield the greatest reduction in human effort while still allowing for dynamic human engagement during generation. Second, while our work has demonstrated the efficacy and potential of knowledge-grounded generation within the Wiki domain—spanning diverse topics such as sports, politics, and science—it is worthwhile to investigate the generalization and adaptability of this method to other, more niche domains, such as law or medicine. These specialized areas may introduce unique challenges due to their inherent complexity. Tabular knowledge may not be readily available or meaningful in such settings, however, it is likely there are other forms of structured knowledge representations that can be consumed in a similar manner. Understanding which prompt and pipeline design choices are best-suited for such targeted domains remains a promising avenue for future research.

\bibliography{acl_latex}

\appendix
\label{sec:appendix}

\section{Multi-document Reasoning Skills Demonstrations}
\label{sec:appendix_reasoning_demonstrations}
\begin{figure*}[t]
\centering
     \includegraphics[width=\textwidth]{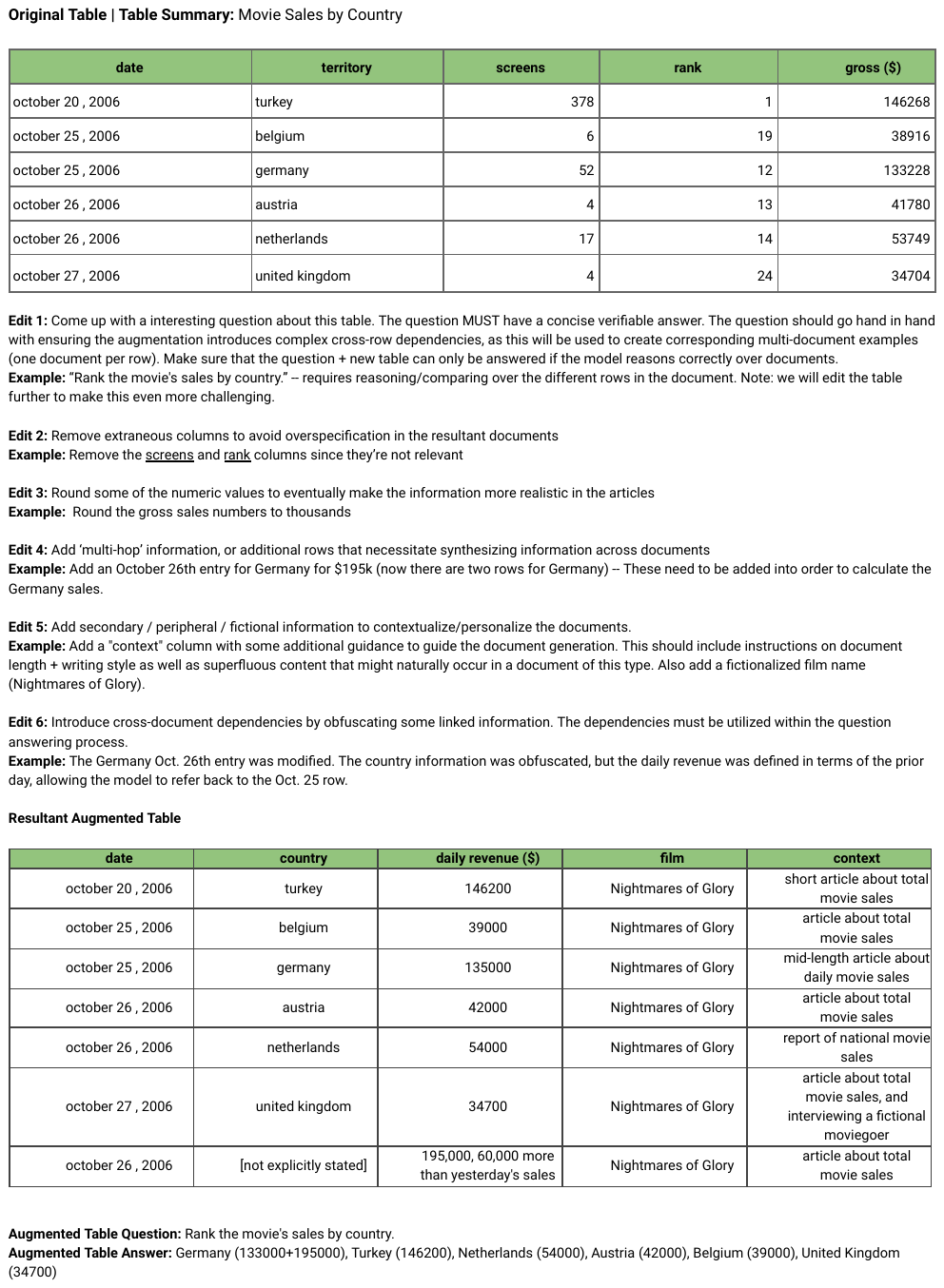}
      \caption{Demonstration of table edit plan used during the knowledge augmentation component of the \MDName{} pipeline.}
     \label{fig:edit_plan_ex}
\end{figure*}

\begin{figure*}[t]
\centering
     \includegraphics[width=0.8\textwidth]{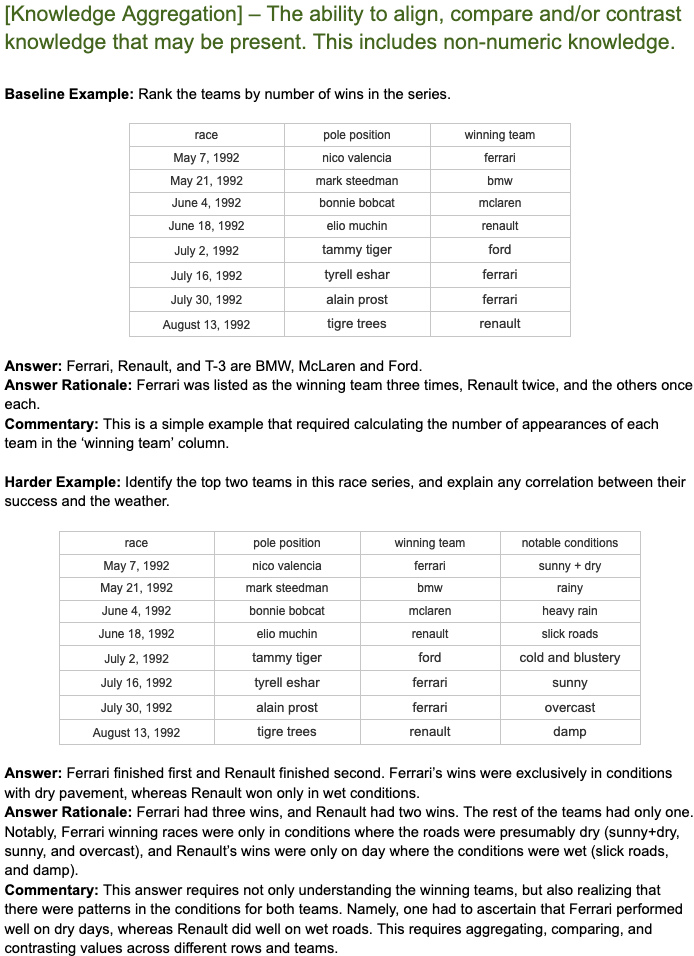}
      \caption{Knowledge Aggregation Skill Description}
     \label{fig:ka_description}
\end{figure*}

\begin{figure*}[t]
\centering
     \includegraphics[width=0.8\textwidth]{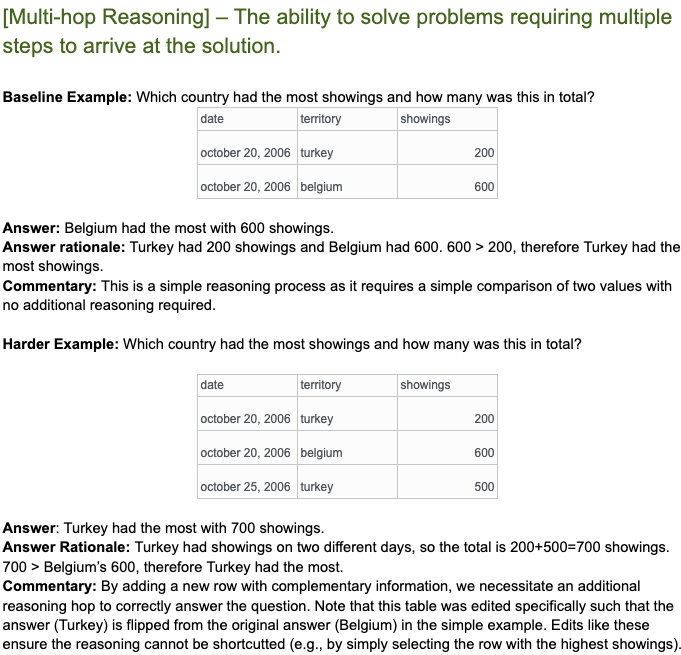}
      \caption{Multi-hop Reasoning Skill Description}
     \label{fig:mh_description}
\end{figure*}

\begin{figure*}[t]
\centering
     \includegraphics[width=0.8\textwidth]{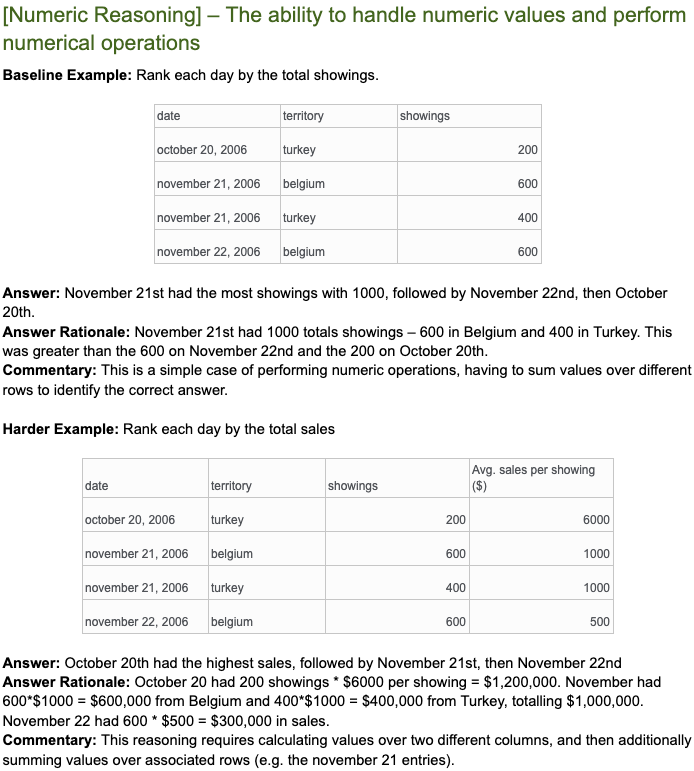}
      \caption{Numeric Reasoning Skill Description}
     \label{fig:numeric_description}
\end{figure*}

\begin{figure*}[t]
\centering
     \includegraphics[width=0.8\textwidth]{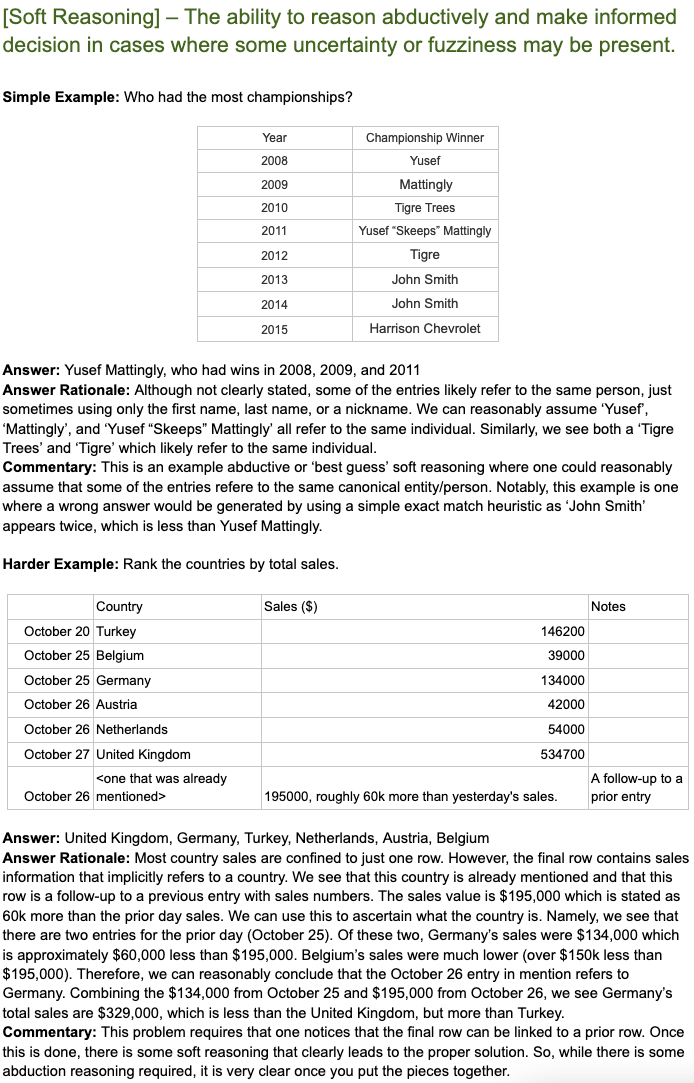}
      \caption{Soft Reasoning Skill Description}
     \label{fig:softr_description}
\end{figure*}

\begin{figure*}[t]
\centering
     \includegraphics[width=0.8\textwidth]{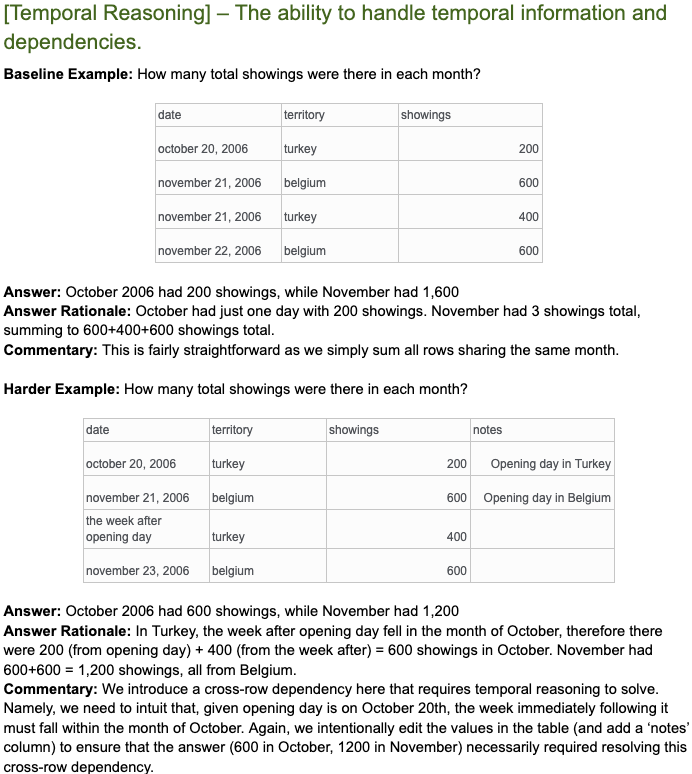}
      \caption{Temporal Reasoning Skill Description}
     \label{fig:temporal_description}
\end{figure*}

Figures \ref{fig:ka_description}, \ref{fig:mh_description}, \ref{fig:numeric_description}, \ref{fig:softr_description}, \ref{fig:temporal_description} overview the five reasoning skills we demonstrate during the creation of \MDName{}.
Figure \ref{fig:edit_plan_ex} demonstrates an edit plan provided to inspire the table augmentation.

\section{Model Evaluation Prompts}
\label{sec:appendix_llm_prompts}

\begin{tcolorbox}[colframe=gray, colback=white, title=Simple QA Prompt]
"You will be presented with a question and a context. You should answer the question based on the context. The last thing you generate should be ANSWER:[your answer here]"
\end{tcolorbox}

\begin{tcolorbox}[colframe=gray, colback=white, title=Chain-of-thought QA Prompt]
"You will be presented with a question and a context. You should answer the question based on the context. Explain your reasoning step by step before you answer. The last thing you generate should be ANSWER:[your answer here]"
\end{tcolorbox}

\section{Example Post-processing and Scoring Prompts}
\label{sec:appendix_postproc_prompts}

These prompts are applied after example construction to
(a) normalise the ground-truth format, (b) extract concise model
predictions after inference, and (c) assign a scalar correctness score to the processed predictions.

\begin{tcolorbox}[colframe=gray,colback=white,title=QA Example Formatting Prompt]
"Adjust the question and answer format such that the question includes
output formatting instructions like ``output the answer as a JSON with
fields \textless{}fields\textgreater{}'', and ensure the answer follows a corresponding JSON
structure.\\
Question: \{question\}\\
Answer: \{answer\}"
\end{tcolorbox}

\begin{tcolorbox}[colframe=gray,colback=white,title=Predicted Answer Extraction Prompt]
"Take this response (which may include verbose reasoning) and concisely
extract only the final answer to the provided question.\\
Question: \{question\}\\
Answer: \{model\_response\}"
\end{tcolorbox}

\begin{tcolorbox}[colframe=gray,colback=white,title=Scoring Prompt]
"On a scale of 0–10, how consistent is the predicted answer with the
ground-truth? Partial credit is acceptable if a portion of the response
matches the ground-truth. RETURN ONLY AN INTEGER.\\
Question: \{question\}\\
Predicted Answer: \{predicted\_answer\}\\
Ground-truth Answer: \{ground\_truth\}"
\end{tcolorbox}

\section{\MDName{} Pipeline Validity Prompts}
\label{sec:appendix_validity_prompts}
We use the following prompts during the knowledge augmentation step to validate the edit plan execution and resultant QA example. Prompt 1 works through the generated problem (leveraging the full knowledge augmentation history) and attempts to rationalize the QA example. Then, prompt 2 evaluates whether this rationalization from Prompt 1 is valid and generates a 0-5 validity scalar, discarding values below 5.

\begin{tcolorbox}[colframe=gray, colback=white, title=Validity Prompt 1]
Original Table Name: \{table\_title\}\\
Original Table: \{original\_table\}\\
Table Edits Applied: \{edits\_applied\}\\
Resultant Table: \{generated\_table\}\\
Resultant Question: \{generated\_question\}\\
Resultant Answer: \{generated\_answer\}\\

Prompt: I have provided an original table, and then an updated version (using the provided knowledge edits) which resulted in an augmented table with a corresponding new question and answer. Use this context and think step by step to come up with a solution rationale that provides a justification for the answer. Note that the original table + edits are provided mostly for added reference. Output the rationale as a string.
\end{tcolorbox}

\begin{tcolorbox}[colframe=gray, colback=white, title=Validity Prompt 2]
How consistent/valid is this reasoning in the following process for generating an example from a table? Score the validity and consistency of the resultant table+question+answer on a scale of 0-5. I want to be able to identify and ignore examples with low scores that I shouldn't include in my dataset.  Output as a json with 'score' and 'explanation' fields. Here is the example: \{prompt\_1\_output\}
\end{tcolorbox}

\begin{figure*}[t]
\centering
     \includegraphics[width=\textwidth]{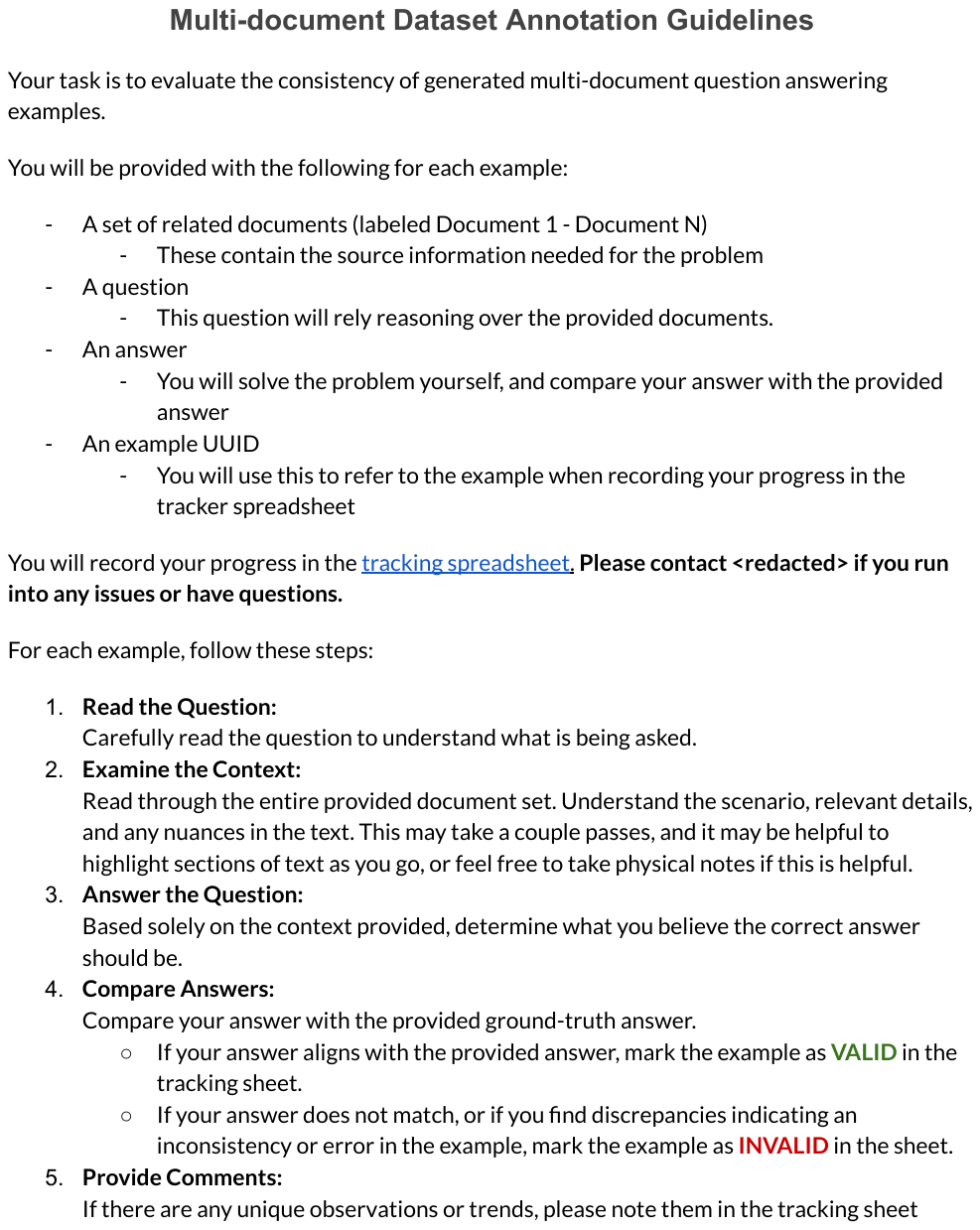}
      \caption{MDBench human validation guidelines.}
     \label{fig:annotation_guidelines}
\end{figure*}

\section{Characteristic Breakdown}
Table \ref{tab:characteristic_analysis} overviews the overall model performance when binning examples by difficulty for each of the five considered characteristics.

\begin{table*}[]
    \centering
    \small
    \begin{tabular}{l|rr|rr|rr|rr|rr}
\hline
 & \multicolumn{2}{c|}{\textbf{Aggregation}}
 & \multicolumn{2}{c|}{\textbf{Multi\text{-}hop}}
 & \multicolumn{2}{c|}{\textbf{Numeric}}
 & \multicolumn{2}{c|}{\textbf{Soft}}
 & \multicolumn{2}{c}{\textbf{Temporal}} \\ 
\textbf{Model} & \textbf{E} & \textbf{H} & \textbf{E} & \textbf{H} & \textbf{E} & \textbf{H} & \textbf{E} & \textbf{H} & \textbf{E} & \textbf{H} \\ \hline
Support                       & 170  & 130  &  89  & 211  & 124  & 176  & 255  &  45  & 208  &  92 \\ \hline
LLaMA\text{-}3\text{-}8B\text{-}Instruct   & 68.0 & 62.6 & 73.6 & 62.0 & 67.1 & 64.8 & 64.4 & 77.0 & 71.5 & 54.0 \\
LLaMA\text{-}3\text{-}70B\text{-}Instruct  & 77.3 & 79.3 & 82.5 & 75.9 & 78.6 & 78.0 & 76.3 & 90.9 & 79.5 & 75.2 \\
Claude\text{-}3\text{-}Opus               & 79.3 & 79.4 & 80.8 & 78.5 & 81.6 & 78.0 & 78.1 & 88.5 & 82.4 & 73.5 \\
Claude\text{-}3.5\text{-}Sonnet           & 79.3 & 76.0 & 83.3 & 75.7 & 85.1 & 73.7 & 77.3 & 84.8 & 81.5 & 70.6 \\
Gemini\text{-}2.5\text{-}Flash            & 74.2 & 82.4 & 76.4 & 78.8 & 80.8 & 74.2 & 78.7 & 72.5 & 80.0 & 68.4 \\
Gemini\text{-}1.5\text{-}Pro              & 78.2 & 75.6 & 83.9 & 74.0 & 85.4 & 72.3 & 75.3 & 88.8 & 81.1 & 68.9 \\
GPT\text{-}3.5\text{-}Turbo               & 68.6 & 72.9 & 74.4 & 68.6 & 72.6 & 69.2 & 69.5 & 75.9 & 74.0 & 62.9 \\
GPT\text{-}4o                             & 79.1 & 79.0 & 83.5 & 77.0 & 84.0 & 76.3 & 79.1 & 78.7 & 83.9 & 69.0 \\
GPT\text{-}o1                             & 79.6 & 82.9 & 80.7 & 81.3 & 80.3 & 82.2 & 81.2 & 80.4 & 81.9 & 77.4 \\ \hline
\textbf{Overall}               & 76.0 & 76.7 & 79.9 & 74.6 & 79.5 & 74.3 & 75.5 & 81.9 & 79.5 & 68.9 \\ \hline
\end{tabular}

    \caption{Characteristic-level Performance Breakdown. We report overall accuracy.}
    \label{tab:characteristic_analysis}
\end{table*}

\newpage

\begin{figure*}[t]
\centering
     \includegraphics[width=\textwidth]{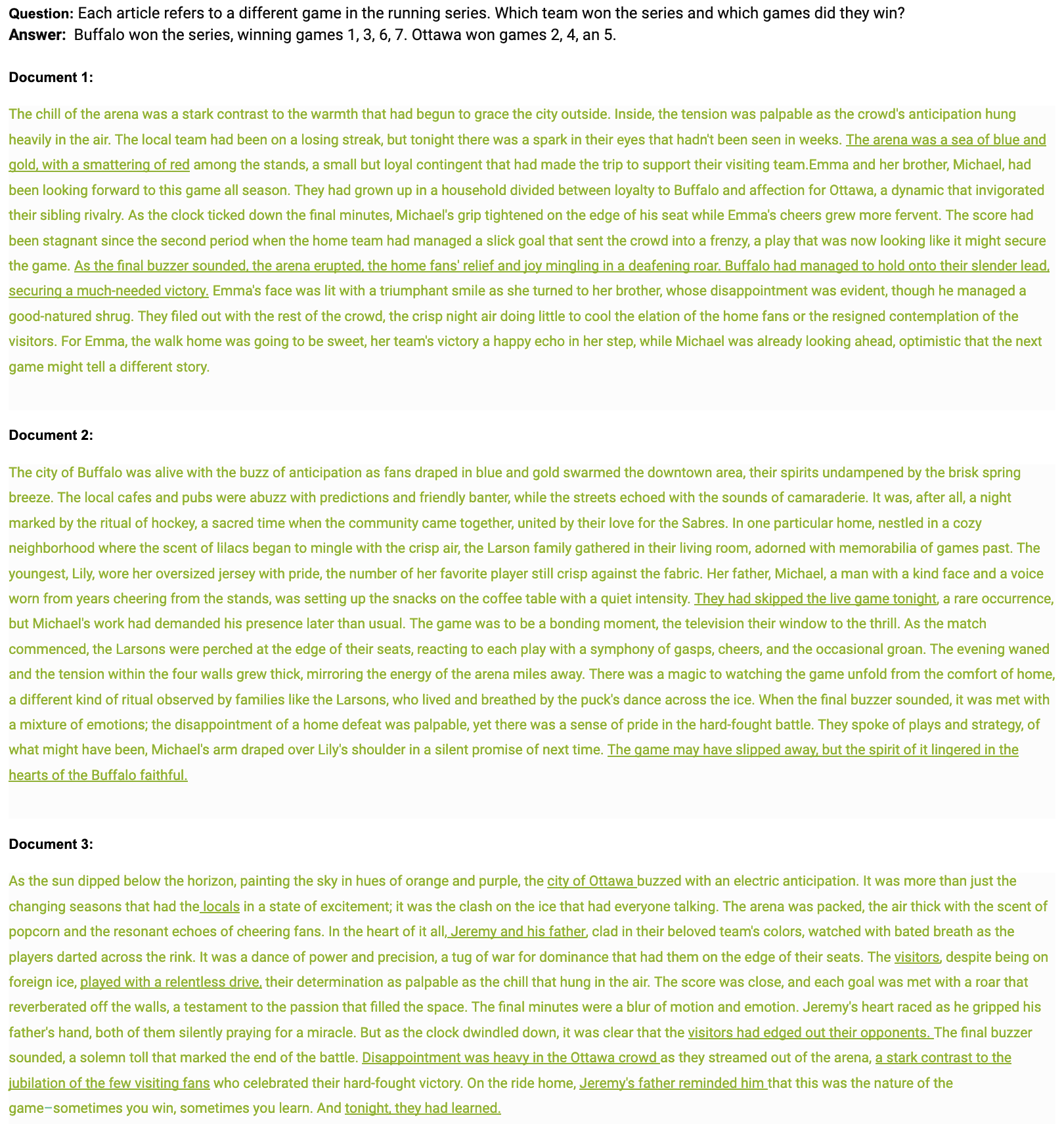}
      \caption{\MDName{} Document QA Example -- Part 1}
     \label{fig:example_1}
\end{figure*}

\begin{figure*}[t]
\centering
     \includegraphics[width=\textwidth]{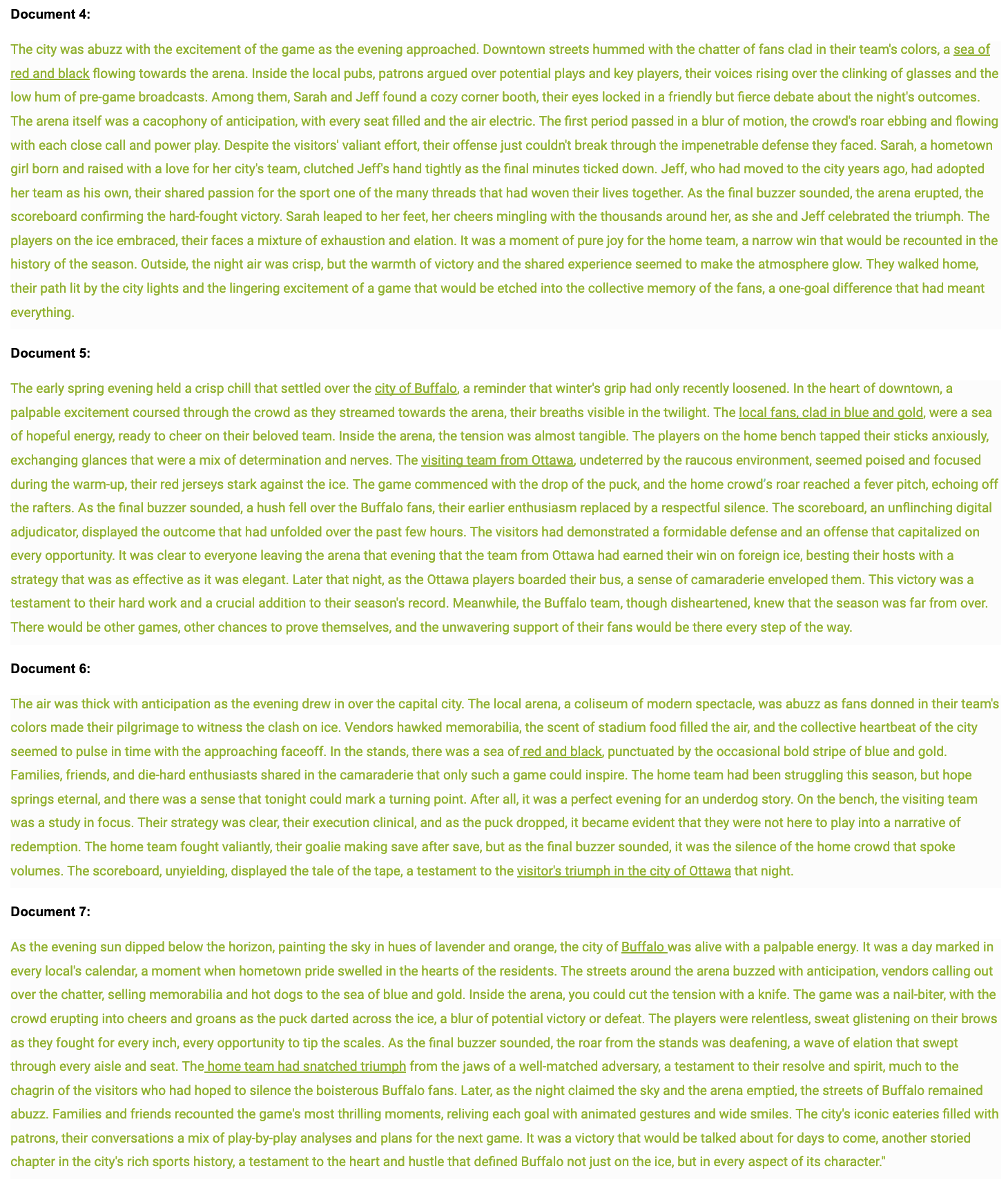}
      \caption{\MDName{} Document QA Example -- Part 2}
     \label{fig:example_2}
\end{figure*}

\end{document}